\documentclass[sigconf]{acmart}

\usepackage{booktabs}
\usepackage{amsmath}

\DeclareMathOperator*{\argmax}{argmax}
\newcommand{\lstm}{\ensuremath{\text{lstm}}}
\newcommand{\tlstm}{\ensuremath{\text{tlstm}}}
\newcommand{\step}{\ensuremath{\text{step}}}

\AtBeginDocument{%
  \providecommand\BibTeX{{%
    \normalfont B\kern-0.5em{\scshape i\kern-0.25em b}\kern-0.8em\TeX}}}

\copyrightyear{2019}
\acmYear{2019}
\setcopyright{acmlicensed}
\acmConference[SIGIR '19]{SIGIR 2019 Workshop on Conversational Interaction Systems.}{July 21--25, 2019}{Paris, France}

\makeatletter
\newcommand*\bigcdot{\mathpalette\bigcdot@{.5}}
\newcommand*\bigcdot@[2]{\mathbin{\vcenter{\hbox{\scalebox{#2}{$\m@th#1\bullet$}}}}}
\makeatother

\begin{document}

%
\title{Dialogue Act Classification in Group Chats with DAG-LSTMs}

%

\author{Ozan \.Irsoy, Rakesh Gosangi, Haimin Zhang, Mu-Hsin Wei, Peter Lund, Duccio Pappadopulo, Brendan Fahy, Neophytos Nephytou, Camilo Ortiz}
\affiliation{Bloomberg L.P.}
\email{ \string{oirsoy, rgosangi, hzhang449, mwei27, plund10, dpappadopulo, bfahy2, nneophytou, cortiz31\string}@bloomberg.net}

%
\renewcommand{\shortauthors}{\.Irsoy et al.}

%
\begin{abstract}
Dialogue act (DA) classification has been studied for the past two decades and has several key applications such as workflow automation and conversation analytics. Researchers have used, to address this problem, various traditional machine learning models, and more recently deep neural network models such as hierarchical convolutional neural networks (CNNs) and long short-term memory (LSTM) networks. In this paper, we introduce a new model architecture, \emph{directed-acyclic-graph LSTM} (DAG-LSTM) for DA classification. A DAG-LSTM exploits the turn-taking structure naturally present in a multi-party conversation, and encodes this relation in its model structure. Using the STAC corpus, we show that the proposed method performs roughly 0.8\% better in accuracy and 1.2\% better in macro-F1 score when compared to existing methods. The proposed method is generic and not limited to conversation applications.
\end{abstract}

%
%

%

%
\maketitle

\section{Introduction}
A dialogue act (DA) is defined as the function of a speaker's utterance during a conversation \cite{mctear2016conversational}, for example, question, answer, request, suggestion, etc. The last two decades have seen many developments in automatic classification of DAs both in spoken \cite{stolcke2000dialogue, ang2005automatic}, and written \cite{wu2005posting, kim2010classifying} conversations. With the increased use of instant messaging and group chat applications, written conversations have become highly prevalent in the modern social and business world. Accurate identification of DAs, especially in business group chats, has many applications such as conversation summarization, question answering, and workflow automation (e.g. reservation systems, scheduling assistants). It is also a critical component in building end-to-end conversational systems \cite{zhao2016towards, young2013pomdp, wen2015semantically}. However, DA classification in written conversations comes with many interesting challenges. Group chats may contain multiple parties conversing simultaneously which leads to entanglement of utterances. Namely, a given utterance could be responding to an utterance that is many turns above or could be starting a brand new conversational thread. Unlike spoken conversations, written conversations do not have any prosodic cues, which have been shown to be useful for DA modeling \cite{fernandez2002dialog, sridhar2009combining}. Due to informal nature of group chats, they tend to contain domain-specific jargon, abbreviations, and emoticons, which further adds to modeling challenges.

Several classical machine learning techniques have been applied to DA classification \cite{ang2005automatic, surendran2006dialog, ribeiro2015influence}. More recently, with the advances in neural networks, deep learning architectures such as Convolutional Neural Networks (CNN), Recurrent Neural Networks (RNN), and their hierarchical variants have also been used for this problem \cite{barahona2016exploiting, ribeiro2018deep, kalchbrenner2013recurrent, liu2017using}. Typically, in these models, the DA of a given utterance is predicted based on three factors: (1) textual content of the utterance, (2) user turns, and (3) contextual information. User turns are usually captured as simple binary features: if the current utterance is from the same user as the prior utterance. Context is obtained either from surrounding utterances within a pre-defined neighborhood window (e.g. the prior two utterances), or from the entire dialog history where influence reduces with distance.

In group chats that have many entangled conversational threads, utterances from a fixed context window might not contain pertinent or sufficient information for DA classification. Likewise, representing context as a flat sequence of all prior utterances would not capture user information: which utterance was posted by which user. This creates a need for a more systematic way to incorporate contextual information for DA classification. To address this issue, we introduce DAG-LSTM, based on tree-LSTMs \cite{tai2015improved}, which are a generalization of LSTMs that support richer network topologies by allowing each LSTM unit to incorporate information from multiple parent units. However, multiple parent units in an LSTM can lead to state explosions because the number of additive terms increase exponentially with the length of the conversation. To this end, we modify the memory cell operations to choose an elementwise maximum over multiple vectors thus effectively choosing a path through one of the child units. We exploit this model architecture to integrate more relevant contextual information for DA classification, and we show that the proposed approach performs better compared to regular LSTMs, CNNs and their hierarchical variants.

The rest of the paper is organized as follows. In the next section, we discuss some historical and relevant work in DA modeling. In Section \ref{methods}, we formulate DA classification as a DAG-LSTM and describe all involved components. This is followed by our experimental work and results. We wrap up the paper with our conclusions and directions for future work.

\section{Background}
Dialogue acts have been studied in linguistics from as early as the 1960s \cite{austin1975things, searle1969speech}. They have become part of computational linguistics \cite{bunt2012iso} in the last two decades especially with the availability of annotated corpora such as the Switchboard corpus \cite{jurafsky1997automatic} and the Meeting Recorder Dialogue Act (MRDA) corpus \cite{shriberg2004icsi}. The Switchboard corpus contains utterances from over 1,155 one-on-one telephonic conversations annotated into 42 different DAs. The MRDA corpus has 75 multi-party meetings labeled into over 50 different DAs. Researchers have used many different machine learning algorithms for DA classification such as Hidden Markov Models (HMM) \cite{surendran2006dialog}, Support Vector Machines (SVM) \cite{ribeiro2015influence}, Maxent classifiers \cite{ang2005automatic}, and Dynamic Bayesian Networks (DBN) \cite{dielmann2008recognition}. 

Kalchbrenner et al. \cite{kalchbrenner2013recurrent} was one of the first to apply deep learning approaches for DA classification, where they used Recurrent Convolutional Neural Networks. Barahona et al. \cite{barahona2016exploiting} used a combination of CNNs for sentence representation and LSTMs for context representation. Both Liu et al. \cite{liu2017using} and Ribeiro et al. \cite{ribeiro2018deep} have studied various combinations of CNNs, LSTMs, and BiLSTMs for sentence and contextual representation.   

Though the majority of research in DA classification has focused on spoken conversations, some recent works have started building DA datasets for written conversations. Kim et al. \cite{kim2012classifying} have published a corpus with 33 live online discussion annotated under 14 dialogue acts. Likewise, Forsyth et al. \cite{forsythand2007lexical} published the NPS chat corpus which contains over 10000 annotated posts that were gathered from various online chat services. More recently, Asher et al. \cite{Asher2016DiscourseSA} published the STAC corpus which includes strategic chat conversations from an online version of the game \textit{The Settlers of Catan}. Another relevant corpus is DailyDialog \cite{li2017dailydialog} which contains one-on-one conversations annotated for emotions, topics, and DAs. For a thorough overview of various dialog-system related corpora, please refer to the review paper by Serban et al. \cite{serban2018survey}. 

\section{Methods}
\label{methods}

\subsection{Problem Statement}
Let $U$ be a group chat consisting of a sequence of utterances $\{u_k\}_{k=1}^K$, where an utterance $u_k$ is written by one of the participants $p_k \in P$ where $P$ denotes the set of chat participants. Given this sequence, the DA classification problem is to assign each utterance $u_k$ a label $y_k \in Y$ where $Y$ denotes the pre-defined dialogue acts.

\subsection{Formulation}
\label{sec:formulation}
We formulate the above stated problem as a sequence modeling task solved specifically using a variant of Tree-LSTMs as described below. Let utterance $u_k$ contain a sequence of words $\{w_t^k\}_{t}$ (which is a shorthand 
for $\{w_t^k\}_{t=1}^T$).
We first map each word $w^k_t$ to a dense fixed-size word vector $\omega^k_t$. Then, an utterance model is used to compute a vector representation $\upsilon_k$ for the entire utterance $u_k$, given its word vector sequence
$\{\omega^k_t\}_t$. Subsequently, a conversation model is used to compute another vector representation
$\phi_k$, given all of the previous utterance vectors $\{\upsilon_j\}_{j=1}^k$, which contextualizes $u_k$ and summarizes the state of the conversation so far. Finally, a classifier layer is used to map $\phi_k$ to $y_k$.
\begin{align*}
\omega^k_t &= \textbf{WordLookup}(w^k_t)\\
\upsilon_k &= \textbf{UtteranceModel}(\{\omega^k_t\}_t)\\
\phi_k &= \textbf{ConversationModel}(\{\upsilon_j\}_{j=1}^k)\\
y_k &= \textbf{Classifier}(\phi_k)
\end{align*}

Below, we describe each of the components in further detail.

\begin{figure}[h]
  \centering
  \includegraphics[width=\linewidth]{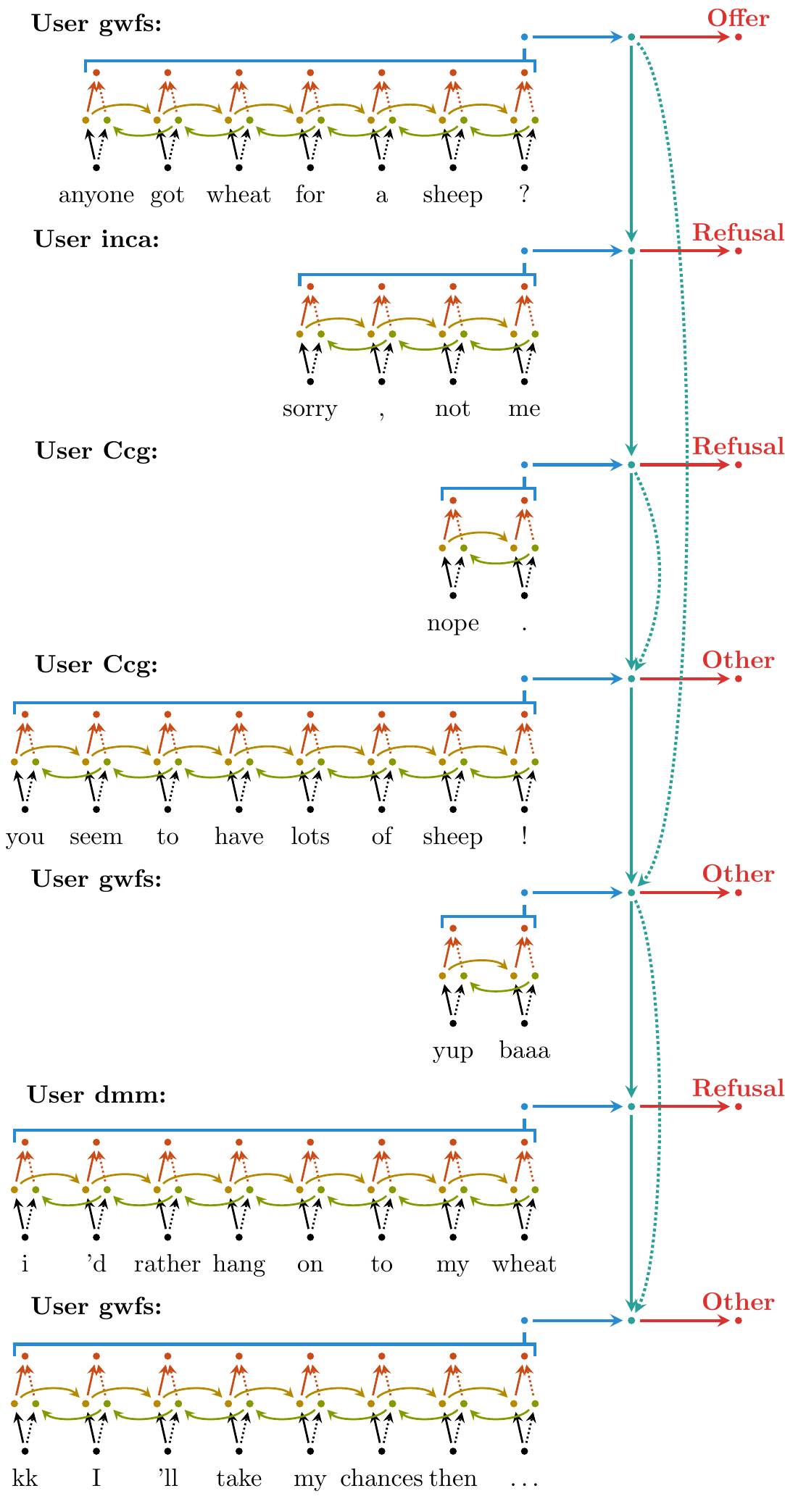}
  \caption{Overall architecture over an example utterance sequence. Arrow color-style combination encodes shared connections within the model. Note the skip connections between consecutive utterances from the same participant, which are added in the form of additional children.}
\end{figure}

\textbf{Representing Utterances.} We use a bidirectional LSTM to represent each 
utterance~\cite{graves2005framewise,hochreiter1997long}.
Let $\lstm(\{x_j\}_{j=1}^t)$ be recursively defined as follows:
\begin{align}
\lstm(\{x_j\}_{j=1}^t) &= \step^{\lstm}\big(x_t, \lstm(\{x_j\}_{j=1}^{t-1})\big)\\
(h_t, c_t) &= \step^{\lstm}\big(x_t, (h_{t-1}, c_{t-1})\big)
\end{align}
where the step function is defined such that
\begin{align}
  i_t &= \text{sigmoid}(W_{ix} x_t + W_{ih} h_{t-1} + W_{ic} c_{t-1} + b_i)\\
  f_t &= \text{sigmoid}(W_{fx} x_t + W_{fh} h_{t-1} + W_{fc} c_{t-1} + b_f)\\
  o_t &= \text{sigmoid}(W_{ox} x_t + W_{oh} h_{t-1} + W_{oc} c_t + b_o) \\
  g_t &= \text{tanh}(W_{cx} x_t + W_{ch} h_{t-1} + b_c)\\
  c_t &= f_t \odot c_{t-1} + i_t \odot g_t\\
  h_t &= o_t \odot \text{tanh}(c_t)
\end{align}
where $W_{\bigcdot}$ and $b_{\bigcdot}$ are weight matrices and bias vectors, respectively,
$i_t$, $f_t$, $o_t$ are \emph{input}, \emph{forget}, and \emph{output} gates, respectively, and
$\odot$ denotes elementwise product.

When the recurrence is defined in terms of the past as above, we get a forward directed LSTM, denoted $\overrightarrow{\lstm}$. Alternatively, we can define it in terms of the future to get a backward
directed LSTM:
\begin{align}
\overleftarrow{\lstm}(\{x_j\}_{j=t}^T) &= \step^{\overleftarrow{\lstm}}\big(x_t, \overleftarrow{\lstm}(\{x_j\}_{j=t+1}^{T})\big)\\
(\overleftarrow{h_t}, \overleftarrow{\vphantom{h}c_t}) &= \step^{\overleftarrow{\lstm}}\big(x_t, (\overleftarrow{h_{t+1}}, \overleftarrow{\vphantom{h}c_{t+1}})\big)
\end{align}

Concatenating $\overrightarrow{h_t}$ and $\overleftarrow{h_t}$, we get a contextualized representation of word
$w_t$ inside the utterance:
\begin{align}
  (\overrightarrow{h_t}, \overrightarrow{c\vphantom{h}_t}) &= \overrightarrow{\lstm}(\{\omega_j\}_{j=1}^t)\\
  (\overleftarrow{h_t}, \overleftarrow{c\vphantom{h}_t}) &= \overleftarrow{\lstm}(\{\omega_j\}_{j=t}^T)\\
\overleftrightarrow{h_t} &= [\overrightarrow{h_t}; \overleftarrow{h_t}]
\end{align}

Finally, contextualized representations are (affinely) transformed into a feature space,
which is then pooled
across all the words in the utterance:
\begin{align}
{\widetilde{\upsilon_t}} &= W_u \overleftrightarrow{h_t} + b_u\\
\upsilon &= \max_t \widetilde{\upsilon_t}
\end{align}
where $\max$ denotes the \emph{elementwise} maximum across multiple vectors and $W_u$ and $b_u$
are the weight matrix and a bias vector, respectively. At the end of this
operation, we have a single fixed size vector $\upsilon$ that represents an entire utterance $u = \{w_t\}_{t=1}^T$.

\textbf{Representing sequences of utterances.}
Given a sequence $\{\upsilon_k\}_k$ of utterance vectors,
a simple method to represent it would be to use another LSTM
model and feed the contextualized (given the history of past utterances) utterance vector to a final
classifier layer:
\begin{align}
(\phi_k, \gamma_k) &= \lstm^{\upsilon}(\{\upsilon_i\}_{i=1}^k)\\
 \hat{y}_k &= \text{softmax}(W_y \phi_k + b_y) \\
 y_k &= \argmax\hat{y}_k
\end{align}
where $W_y$ and $b_y$ is a weight matrix and vector, and $\hat{y}_k$ denotes the predicted probability
distribution over the dialogue act set $Y$.

In this approach, a conversation would be represented as a flat sequence of utterances with no information about which utterance belongs to which participant. In order to address this, we add skip connections between consecutive posts \emph{from the same participant}. This means that each utterance has two antecedents: (1) past utterance and (2) past utterance from the same participant. Doing so, we achieve
two things: the model can build up a user history and link each utterance to a user's particular history within conversation, and also make utterances from the same user closer in the computation graph.

To this end, we employ \textbf{Tree-LSTM} equations which were previously used for similar computation graphs 
where each node in a graph has more than one child~\cite{tai2015improved}.

Let $\tlstm(\{x_{\eta'}\}_{\eta'\in\text{Sp}(\eta)}) = (h_\eta, c_\eta)$ denote a Tree-LSTM where
$\eta$ is a node in a given tree or a graph,
$\text{Sp}(\eta)$ denotes the index set for the subtree (subgraph) spanned by $\eta$
and $\{x_{\eta'}\}_{\eta'\in\text{Sp}(\eta)}$ 
denotes the nodes spanned by $\eta$.
Then, $\tlstm$ is recursively defined in terms of the children of $\eta$, $ch(\eta)$:
\begin{align}
\tlstm(\{x_{\eta'}\}_{\eta'\in\text{Sp}(\eta)}) &= \step^{\tlstm}\bigg(x_\eta, \bigcup_{\eta' \in ch(\eta)} \tlstm(\{x_{\eta''}\}_{\eta'' \in \text{Sp}(\eta')})\bigg)\\
(h_\eta, c_\eta) &= \step^{\tlstm}\bigg(x_\eta, \bigcup_{\eta' \in ch(\eta)} (h_{\eta'}, c_{\eta'}) \bigg)
\end{align}
where the step function is defined such that:
\begin{align}
    i_\eta &= \text{sigmoid}(W_{ix} x_\eta + \sum_{\eta'\in ch(\eta)} W_{ih}^{e(\eta',\eta)}h_{\eta'} + b_i)\\
    f_{\eta\eta'} &= \text{sigmoid}(W_{fx} x_\eta + \sum_{\eta''\in ch(\eta)} W_{fh}^{e(\eta',\eta)e(\eta'',\eta)}h_{\eta''} + b_f)\\
    o_\eta &= \text{sigmoid}(W_{ox} x_\eta + \sum_{\eta'\in ch(\eta)} W_{oh}^{e(\eta',\eta)}h_{\eta'} + b_o)\\
    g_\eta &= \text{tanh}(W_{gx} x_\eta + \sum_{\eta'\in ch(\eta)} W_{gh}^{e(\eta',\eta)}h_{\eta'} + b_g)\\
    c_\eta &= i_\eta \odot g_\eta + \sum_{\eta'\in ch(\eta)} f_{\eta\eta'} \odot c_{\eta'} \label{eq:newcell}\\
    h_\eta &= o_\eta \odot \text{tanh}(c_\eta)
\end{align}
where $e(\eta', \eta) \in E$ denotes the edge type (or label) that connects $\eta'$ to $\eta$. In general, $E$ can be an arbitrary fixed size set. In our work it is of size two: edges that connect past utterance to current utterance and edges that connect past utterance from the same participant to the current utterance. Since weights are parametrized by the edge types $e(\eta', \eta)$, contribution of past utterance vs. past utterance from the same participant is computed differently.

Note that we can apply Tree-LSTM equations even though our computation graphs are not trees but directed acyclic graphs (DAGs), since each node feeds into not one parent but two (next utterance and next utterance from the participant). 

A key observation related to this fact is as follows: Let us consider the sink node (last utterance in a conversation) memory cell $c_{sink}$. Since each node cell $c_\eta$ contributes to not one but two other cells $c_{\eta'}$, and $c_{\eta''}$ additively, recursively unfolding Eq.~\ref{eq:newcell} for $c_{sink}$ gives exponentially many additive terms of $c_\eta$ in the length of the shortest path from $\eta$ to $sink$. This causes very quick state explosions in the length of a conversation, which we experimentally confirm.

To combat this, we make a very simple modification to Eq.~\ref{eq:newcell} as follows:
\begin{align}
    c_\eta &= i_\eta \odot g_\eta + \max_{\eta'\in ch(\eta)} f_{\eta\eta'} \odot c_{\eta'}
\end{align}
where $\max$ denotes the elementwise maximum over multiple vectors, which effectively picks (in an elementwise fashion) a path through either one of the children. Thus, cell growth can be at worst linear in the conversation length.

Since the modified equations are more appropriate for DAGs compared to Tree-LSTMs which suffer from explosions, we call the modified model \textbf{DAG-LSTM}. Note that both Tree-LSTM and DAG-LSTM reduces to classical LSTMs (without the peephole connections) when each node has exactly one children and one parent.

On top of the DAG-LSTM, classification layer works same as before:
\begin{align}
  (\phi_k, \gamma_k) &= \text{daglstm}^{\upsilon}(\{\upsilon_i\}_{i=1}^k)\\
  \hat{y}_k &= \text{softmax}(W_y \phi_k + b_y)
\end{align}

\textbf{Related architectures.}
To our knowledge there are two DAG-based variants of the LSTM architecture~\cite{zhu2016dag, chen2017dag},
both of which operate differently than ours. In \cite{zhu2016dag}, nodes with multiple children require
a binarization operation specific to a task, and the contribution of each child is computed using the 
same weights. In \cite{chen2017dag}, the architecture operates more similarly to ours, however the past cell
state of a parent node is defined as a simple sum of all children states, and subsequently, traditional LSTM 
updates are used. Our approach is the most faithful to the original Tree-LSTM updates which have been studied 
before in many applications~\cite{tai2015improved,zhu2015long,chen2016enhancing,maillard2017jointly,chen2018tree}. 
Furthermore, neither of the DAG-based approaches address the inherent problem
of state explosion as described in Section~\ref{sec:formulation}.

\section{Experimental work}

We compare the performance of our proposed model with four baseline architectures that were employed in prior works \cite{barahona2016exploiting, ribeiro2018deep, liu2017using}. The first baseline model uses CNNs for both utterance and context representation. The second baseline model uses BiLSTMs for utterance representation and LSTMs for context representation. The third model employs CNNs for utterance representation and LSTMs for context representation. The last baseline model uses BiLSTMs for utterance representation and has no context representation. Finally, our model uses BiLSTMs for utterance representation and DAG-LSTMs for context representation. We explicitly chose not to use BiLSTMs for context representation because such architectures are not viable for live systems. We evaluated all five models on the STAC corpus.

\textbf{Data.} The STAC corpus \cite{Asher2016DiscourseSA} contains conversations from an online version of the game \textit{The Settlers of Catan}, where trade negotiations were carried out in a chat interface. The data contains over 11000 utterances from 41 games annotated for various tasks such as anaphoric relations, discourse units, and dialog acts. For our experimental work, we only used the dialog act annotations. The corpus had six different DAs but one of those acts named Preference had very low prevalence (only 8 utterances). Therefore, we excluded it from our experimental work. Table~\ref{table:class-dist} below shows utterance counts for all six DAs.

\begin{table}
    \caption{Frequency of dialog acts}
	\label{table:class-dist}
	\begin{tabular}{lr}
		\toprule
		Dialog Act & Count \\
		\midrule
        Accept & 658 \\
        Counteroffer & 643 \\
        Offer & 1571 \\
        Other & 7012 \\
        Refusal & 1546\\
        \textit{Preference (discarded)} & 8 \\
        \bottomrule
	\end{tabular}
\end{table}

We split the data randomly into three groups: train (29 games - 8250 utterances), dev (4 games - 851 utterances), and test (8 games - 2329 utterances). The utterances were tokenized using the Stanford PTBTokenizer \cite{manning2014stanford} and the tokens are represented using GloVe embeddings \cite{pennington2014glove}. 

\begin{table*}[tbh]
	\caption{F1 scores of various classes for different models.}
	\begin{tabular}{lcccccccc}
		\toprule
		Model         &  Accept & Counteroffer & Offer  & Other & Refusal & \hspace{5pt} &Accuracy & Macro-F1  \\ \midrule
		CNN + LSTM    & \textbf{65.64} & 51.06 & 77.54 & 93.57 & 84.33 & &86.43 & 74.43 \\ 
		CNN + CNN     & 65.42 & 47.89 & 75.81 & 92.00 & 83.66 & &85.32 & 73.10 \\
		BiLSTM + no utterance context \hspace{5pt} & 46.83 & 41.24 & 71.64 & 89.01 & 76.28 & & 79.49 & 65.00 \\
		BiLSTM + LSTM & 63.93 & 50.17 & 79.03 & 94.12 & 85.59 & & 86.83 & 74.57 \\ 
		BiLSTM + DAG-LSTM &  64.29 & \textbf{51.69} & \textbf{81.97} & \textbf{94.42} & \textbf{86.54} & & \textbf{87.69} & \textbf{75.78} \\ \bottomrule
	\end{tabular}
	\label{table:results}
\end{table*}

\begin{figure*}
\includegraphics[width=0.49\textwidth]{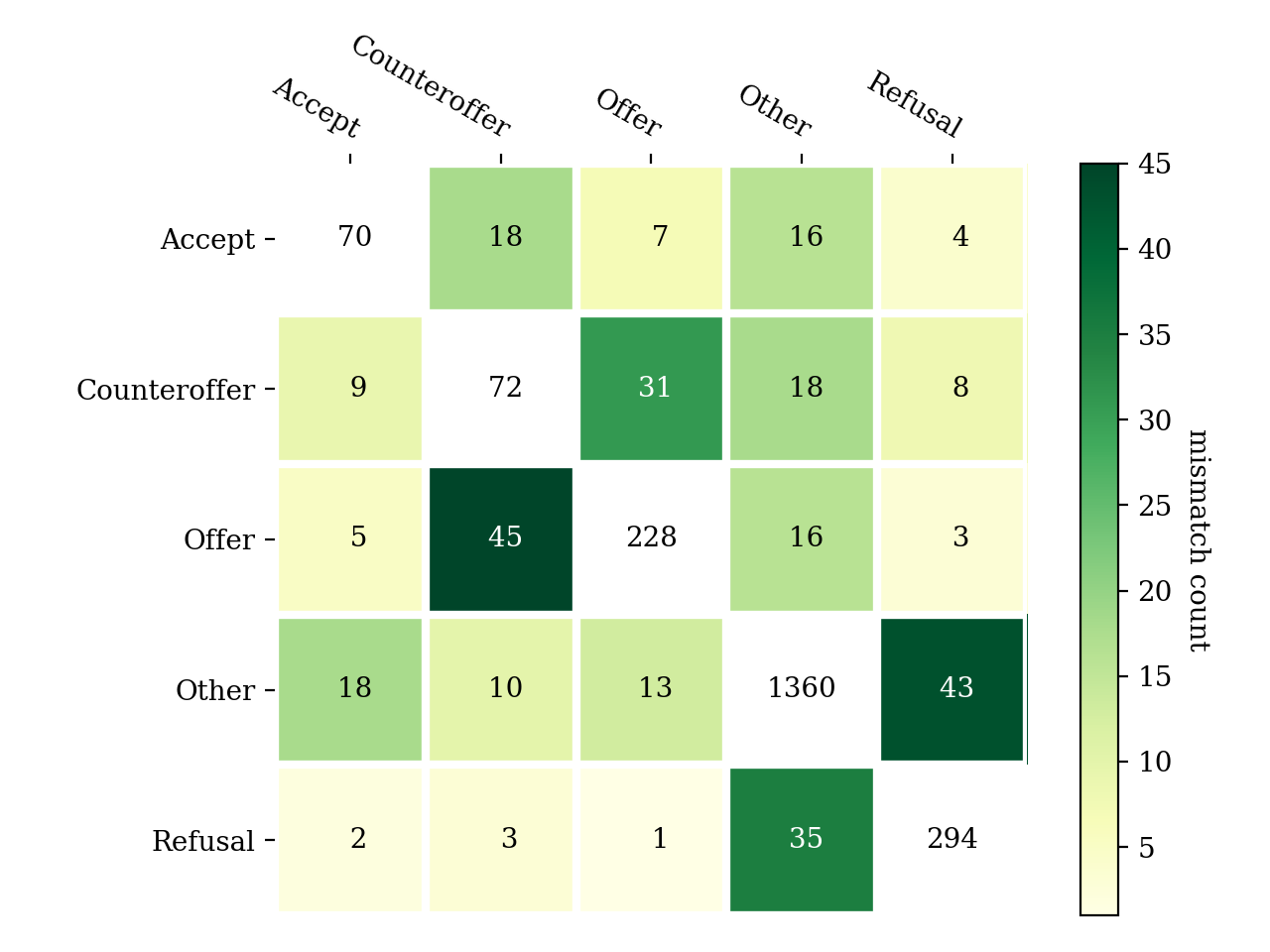}
\includegraphics[width=0.49\textwidth]{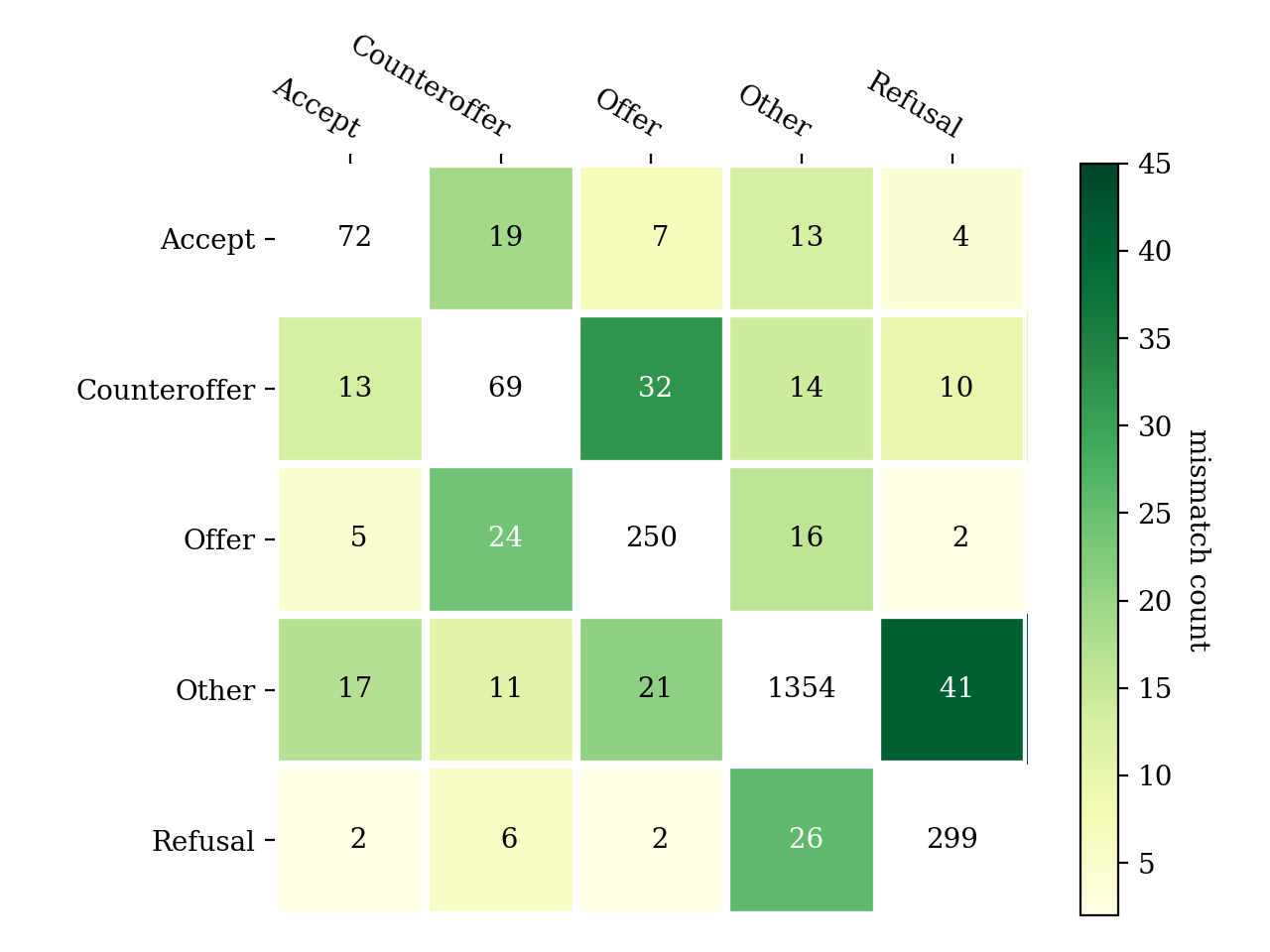}
\caption{Confusion matrices for (left) BiLSTM + LSTM and (right) BiLSTM + DAG-LSTM. Rows denote gold labels whereas columns denote predicted labels by the model.}
\label{fig:conf}
\end{figure*}

\textbf{Setting.} We use Adam optimizer in the stochastic gradient descent setting to train all models~\cite{kingma2014adam}.
We use a patience value of 15 epochs, i.e. training is stopped after not observing an improvement for 15 epochs in the validation data, and train for a maximum of 300 epochs. We pick the best iteration based on the validation macro-F1 score.
All five models have been hyperparameter-tuned using validation set macro-F1 using simple random search. We run a total of 100 experiments to evaluate random hyperparameter candidates based on the following distributions (whenever applicable to a particular architecture): 

\begin{tabular}{ll}
Learning rate & $\sim 10^{\text{Uniform}(-5,-3)}$\\
Dropout rate & $\sim$ Uniform(0, 0.5)\\
Word dropout rate & $\sim$ Uniform(0, 0.3)\\
Word vector update mode & $\sim$ Uniform\{fixed, fine-tune\}\\
\#Units in utterance layer & $\sim$ Uniform\{50, 75, 100, 200\}\\
\#Units in conversation layer & $\sim$ Uniform\{50, 75, 100, 200\}\\
\#Filters in CNNs &  $\sim$ Uniform\{50, 75, 100, 200\} \\
Window size for CNNs &  $\sim$ Uniform\{2,3,4\} \\
\end{tabular}

\section{Results}
Table~\ref{table:results} summarizes the results in terms of F1 scores for individual classes, overall accuracy, and macro-F1 score. The BiLSTM + DAG-LSTM architecture achieves the best F1 score for four classes. The overall accuracy of 87.69\% is 0.86\% better than the second best model (BiLSTM + LSTM). Likewise, the macro-F1 score of 75.78\% is over 1\% better than the next best model.

The owners of STAC corpus have presented results \cite{cadilhac2013grounding} from using CRFs for this problem on a preliminary version of the dataset which contained utterances from only 10 games. Their models are reported to have achieved 83\% accuracy and 73\% macro-F1 score. Though these numbers are not directly comparable with the results in Table~\ref{table:results}, we wanted to present them here for complete context. 

Confusion matrices for BiLSTM + LSTM and BiLSTM + DAG-LSTM are shown in Figure~\ref{fig:conf}. We see that BiLSTM + DAG-LSTM has  less confusion correctly classifying offers, specifically by avoiding mistakenly classifying as Counteroffer. This suggests that  using  additional context information provided by skip connections is helpful, since the utterances for Offer and Counteroffer are typically similar. We also see less confusion misclassifying Refusal as Other.

\begin{table*}[tbh]
     \caption{Errors made by different models.}
     \label{table:errors}
     \begin{tabular}{llllll}
     \toprule
     Participant & Conversation (tokenized) & \hphantom{Hi} & Gold & DAG-LSTM & LSTM \\
     \midrule
     & ... & & & & \\
     william      & can i get a clay from someone ?   && Offer   & Offer   & Offer \\
     ljaybrad123  & none sorry                        && Refusal & Refusal & Refusal \\
     tomas.kostan & anyone have some wood to spare ?  && Offer   & Offer   & Offer \\
     william      & for clay ...                      && Counteroffer & Counteroffer & Counteroffer \\
     ljaybrad123  & no sorry                          && Refusal & Refusal & Refusal \\
     tomas.kostan & for a sheep                       && Offer   & Offer   & {\color{red} Counteroffer} \\
     william      & for ore ?                         && Counteroffer & Counteroffer & Counteroffer \\
     tomas.kostan & can only offer a sheep            && Offer   & Offer   & {\color{red} Counteroffer} \\
     william      & got enough . sorry                && Refusal & Refusal & Refusal \\
     william      & i can give you a lot of clay, anyone ? && Offer   & Offer   & {\color{red} Counteroffer} \\
     tomas.kostan & no sry                            && Refusal & Refusal & Refusal \\
     & ... & & & & \\
     \midrule
     & ... & & & & \\
     tomas.kostan  & no clay ?                        && Offer   & Offer   & Offer \\
     ljaybrad123   & i have clay                      && Accept  & {\color{red}Counteroffer} & {\color{red}Offer} \\
     gotwood4sheep & LJ has one                       && Other   & Other   & Other \\
     ljaybrad123   & I want wood                      && Counteroffer & Counteroffer & {\color{red}Offer} \\
     gotwood4sheep & my one grrrrrr                   && Other   & Other   & {\color{red}Refusal} \\
     ljaybrad123   & :D                               && Other   & Other   & Other \\
     & ... & & & & \\
     \midrule
     & ... & & & & \\
     Kersti        & anyone need sheep?I              && Offer   & Offer   & Offer \\
     Kersti        & I have a large quantity of them  && Offer   & Offer   & {\color{red}Other} \\
     Kersti        & and could do with wheat / wood instead && Offer & Offer & {\color{red}Counteroffer} \\
     Tyrant Lord   & no                               && Refusal & Refusal & Refusal \\
     Tyrant Lord   & sorry                            && Other   & Other   & Other \\
     Kersti        & it 's good quality wool this season ! && Other   & Other   & Other \\
     sparkles      & i do want sheep ,                && Other   & Other   & Other \\
     sparkles      & but I need all my resources .    && Other   & Other   & {\color{red}Refusal} \\
     sparkles      & I might want to trade after I roll ... && Refusal & {\color{red}Other} & {\color{red}Other}\\
     sparkles      & sorry for being rubbish !        && Other   & Other   & Other \\
     & ... & & & & \\
     \bottomrule
     \end{tabular}
\end{table*}

To verify these effects we present some example conversations with predicted outputs from BiLSTM + LSTM and BiLSTM + DAG-LSTM in Table~\ref{table:errors}, to showcase errors made by different models. In these examples, we repeat the observation that the architecture with DAG-LSTM has less confusion between Offer and Counteroffer. We also observe some utterances that both architectures failed to classify correctly: second utterance in second conversation and second to last post in the last conversation.

\section{Conclusion}
In this paper, we introduced a new architecture (DAG-LSTM) that provides a systematic way to incorporate contextual information for DA classification. We evaluated the model on STAC corpus and compared it with hierarchical LSTM and CNN models. Our experimental work shows that the DAG-LSTM achieves much better accuracy and macro-F1 scores compared to state-of-the-art baseline models. In particular, the results demonstrate that information about the prior utterance made by a speaker is very useful in DA classification. This approach facilitates learning relevant context by skipping potentially irrelevant utterances from other speakers in a chat room. We can extend this idea to other types of context, such as the prior utterance from the same team when group membership is available, or prior utterance from the same conversational thread. We propose to experiment with these different types of context in the future.

In this paper, we mainly focused on multi-party written conversations. Despite growing interest in dialog modeling, there are rather very few datasets with DA annotations in the written domain especially in a group-chat setting, excluding transcribed versions of spoken conversations. To the best of our knowledge, there are only two such datasets: the STAC Corpus and the NPS chat corpus. Unfortunately, we could not use the NPS chat corpus because of licensing issues. We expect more such datasets will be made available for research in the future because of the wide-spread usage of group chat applications. 

Though our paper has focused only on DA classification, we expect the architecture presented here could be used to address many other aspects of dialog modeling such as emotion classification, sentiment analysis, and thread disentanglement. Likewise, we expect the methodology presented here could easily be extended to address DA classification in spoken conversations.

\begin{acks}
The authors would like to thank Heidi Johnson, the Community, Collaboration and Compliance Product team, and the Office of the CTO at Bloomberg for their support in the development of this paper.
\end{acks}

%
\bibliographystyle{ACM-Reference-Format}
\bibliography{sample-base}

\end{document}